\documentclass{article}
\usepackage[top=1in, bottom=1in, left=1in, right=1in]{geometry}
\usepackage{tgpagella}
\usepackage[utf8]{inputenc} %
\usepackage[T1]{fontenc}    %

\usepackage{url}
\usepackage[hidelinks]{hyperref}
\usepackage{amsmath,amsthm,amssymb,amsbsy}
\usepackage{paralist}
\usepackage{xcolor}
\usepackage{color}
\usepackage{graphicx}
\graphicspath{{./figs/}}
\usepackage{algorithm}
\usepackage{algorithmic}
\usepackage{comment}
\usepackage{multirow}

\usepackage{graphicx}
\usepackage{fancyhdr}
\usepackage{cite}
\usepackage{cleveref}
\usepackage{subcaption}

\theoremstyle{remark}

\theoremstyle{problem}

\newcommand{\R}{\mathbb{R}}

\newcommand{\e}{\begin{equation}}
\newcommand{\ee}{\end{equation}}
\newcommand{\en}{\begin{equation*}}
\newcommand{\een}{\end{equation*}}
\newcommand{\eqn}{\begin{eqnarray}}
\newcommand{\eeqn}{\end{eqnarray}}
\newcommand{\bmat}{\begin{bmatrix}}
\newcommand{\emat}{\end{bmatrix}}

\DeclareMathAlphabet\mathbfcal{OMS}{cmsy}{b}{n}

\newcommand{\vct}[1]{\boldsymbol{#1}}

\newcommand{\wh}{\widehat}

\newcommand{\innerprod}[2]{\left\langle #1,  #2 \right\rangle}

\newcommand{\calV}{\mathcal{V}}

\newcommand{\vh}{\vct{h}}

\newcommand{\vv}{\vct{v}}

\newcommand{\vmu}{\vct{\mu}}

\graphicspath{{./figs/}}

\newlength{\imgwidth}
\newboolean{twoColVersion}
\setboolean{twoColVersion}{false}
\newcommand{\twoCol}[2]{\ifthenelse{\boolean{twoColVersion}} {#1} {#2} }

\usepackage{amsmath}

\usepackage[utf8]{inputenc} %
\usepackage[T1]{fontenc}    %
\usepackage{wrapfig}
\usepackage{url}            %
\usepackage{booktabs}       %
\usepackage{amsfonts}       %
\usepackage{nicefrac}       %
\usepackage{microtype}      %
\usepackage{xcolor}         %
\usepackage{multirow}
\usepackage{graphicx}       %
\usepackage{xcolor}
\usepackage{enumerate}
\usepackage{enumitem}
\usepackage{authblk}
\usepackage{xspace}

\usepackage{fontawesome5} %

\newcommand{\mean}{\operatorname{mean}}

\newcommand{\pt}{\mbox{$\mathcal{A}_{\mathop{\mathtt{pt}}\limits}$}\xspace}

\newcommand{\ft}{\mbox{$\mathcal{A}_{\mathop{\mathtt{ft}}\limits}$}\xspace}

\newcommand{\model}{\mbox{$\mathcal{A}$}\xspace}

\newcommand{\method}{\mbox{reflection-inducing probing}\xspace}

\newcommand{\reflect}{\mbox{$\mathbb{H}^{(\ell)}_{\mathop{\mathtt{reflect}}\limits}$}\xspace}

\newcommand{\nonreflect}{\mbox{$\mathbb{H}^{(\ell)}_{\mathop{\mathtt{non-reflect}}\limits}$}\xspace}

\newcommand{\refvec}{\mbox{$\vmu^{(\ell)}_{\mathop{\mathtt{reflect}}\limits}$}\xspace}

\newcommand{\nonrefvec}{\mbox{$\vmu^{(\ell)}_{\mathop{\mathtt{non-reflect}}\limits}$}\xspace}

\title{From Emergence to Control: Probing and Modulating Self-Reflection in Language Models}

\author{
Xudong Zhu, Jiachen Jiang, Mohammad Mahdi Khalili, Zhihui Zhu
  \\  \vspace{1ex}
Department of Computer Science \& Engineering, The Ohio State University \\ \vspace{1ex}
  \;\texttt{\{zhu.3944,\;\;jiang.2880,\;\;khaliligarekani.1,\;\;zhu.3440\}@osu.edu} \\
  \faGithub\hspace{0.3em}\href{https://github.com/xzAscC/ProbingReflection}{ProbingReflection}
}

\begin{document}

\maketitle

\begin{abstract}
  Self-reflection---the ability of a large language model (LLM) to revisit, evaluate, and revise its own reasoning---has recently emerged as a powerful behavior enabled by reinforcement learning with verifiable rewards (RLVR). While self-reflection correlates with improved reasoning accuracy, its origin and underlying mechanisms remain poorly understood. In this work, {\it we first show that self-reflection is not exclusive to RLVR fine-tuned models: it already emerges, albeit rarely, in pretrained models}. To probe this latent ability, we introduce Reflection-Inducing Probing, a method that injects reflection-triggering reasoning traces from fine-tuned models into pretrained models. This intervention raises self-reflection frequency of Qwen2.5 from 0.6\% to 18.6\%, revealing a hidden capacity for reflection. Moreover, our analysis of internal representations shows that both pretrained and fine-tuned models maintain hidden states that distinctly separate self-reflective from non-reflective contexts. Leveraging this observation, {\it we then construct a self-reflection vector, a direction in activation space associated with self-reflective reasoning}. By manipulating this vector, we enable bidirectional control over the self-reflective behavior for both pretrained and fine-tuned models. Experiments across multiple reasoning benchmarks show that enhancing these vectors improves reasoning performance by up to 12\%, while suppressing them reduces computational cost, providing a flexible mechanism to navigate the trade-off between reasoning quality and efficiency without requiring additional training. Our findings further our understanding of self-reflection and support a growing body of work showing that understanding model internals can enable precise behavioral control.
\end{abstract}

\section{Introduction}

Reinforcement Learning with Verifiable Rewards (RLVR) has emerged as a powerful technique for enhancing the reasoning abilities of large language models (LLMs), enabling learning from outcome-level feedback across diverse tasks \cite{xu2025towards, wang2025selfreasoning, mroueh2025reinforcement}. In a nutshell, RLVR optimizes for end-task success, allowing models to explore novel reasoning strategies at scale \cite{zhao2025r1, ferrag2025reasoning, su2025expanding}. Notably, it has been reported \cite{guo2025deepseek, liu2025understanding, zeng2025simplerl} that such training induces new emergent behaviors, such as self-reflection—the ability of a model to revisit, evaluate, and revise its prior outputs. For instance, the DeepSeek-R1 report \cite{guo2025deepseek} highlights that RLVR-trained models often generate tokens such as “wait”, interpreted as signals of internal deliberation or critique.

Despite these observations, it remains unclear why such reflective behaviors emerge. Answering this question is crucial for understanding the foundations of reasoning in LLMs and guiding future methods for enhancing their performance. Moreover, empirical studies have shown that reflection correlates with more accurate and robust reasoning \cite{zuo2025ttrl, yue2025does, liu2024self}, and that prompting models to explicitly “wait” or reflect can further improve performance with test-time computing. However, this benefit may come at a cost: reflection can increase inference time, introduce unnecessary verbosity, and reduce computational efficiency \cite{yang2025dynamic, renze2024self, sui2025stop}. These findings highlight not only the need to understand self-reflection in LLMs, but also the importance of controlling it to balance reasoning quality and efficiency. These gaps motivate this work to study the following research questions:
\begin{center}
  \textit{Is self-reflection a novel behavior induced by RLVR, or does it already emerge during pretraining? Can we control self-reflection in LLMs to balance performance and computational efficiency?}
\end{center}

\begin{figure}
  \includegraphics[width=\linewidth]{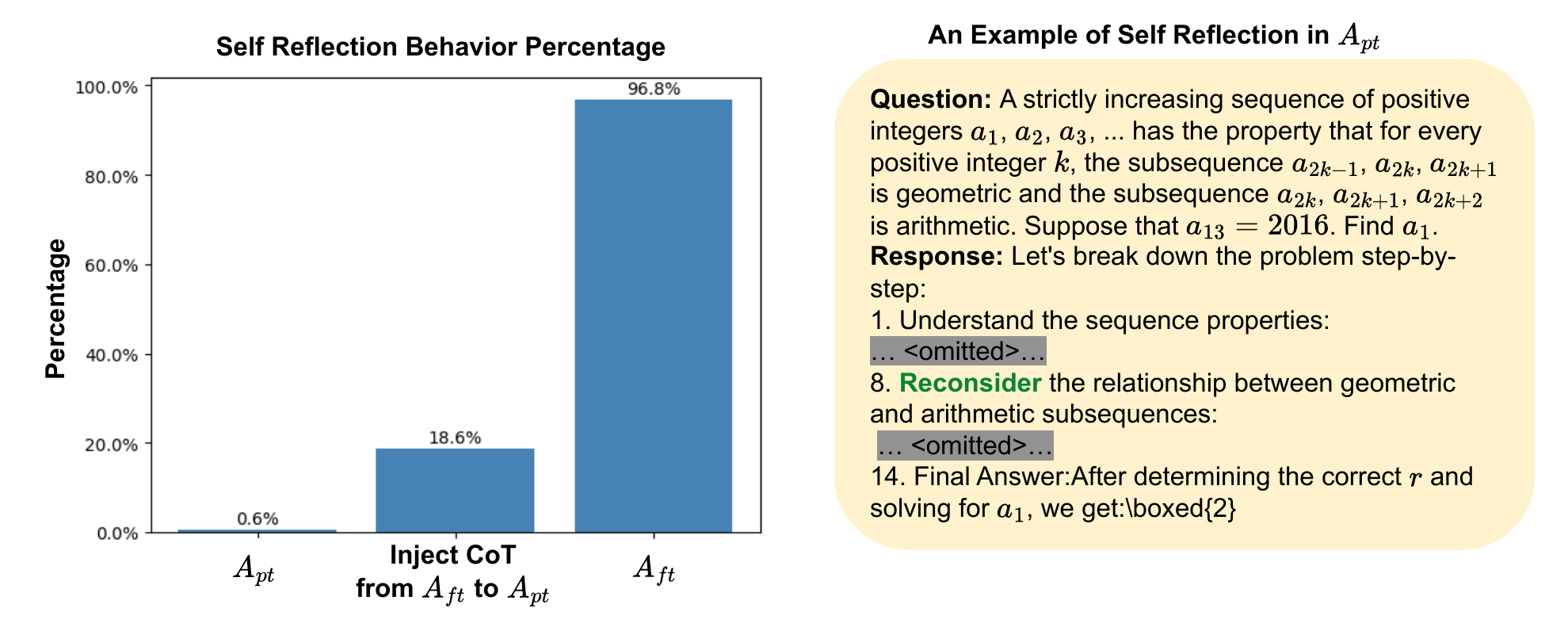}
  \caption{\textbf{Left:} Frequency distribution of self-reflection behaviors for pretrained model \pt, fine-tuned model \ft, and \pt with \method by injecting CoT from \ft, evaluated  on the MATH500 dataset. \textbf{Right:} A representative example of spontaneous self-reflection in \pt, demonstrating that this capability emerges naturally during pretraining, albeit with different self-reflection tokens than those typically observed in RLVR fine-tuned models.}
  \label{fig:token_distribution}
\end{figure}

\paragraph{Contribution}

In this work, we provide affirmative answers to both questions. First, we compare the reasoning behaviors of pretrained models and fine-tuned models (either via RLVR or distillation), and verify that self-reflection is already present in the pretrained model, albeit at a much lower frequency. Next, we analyze the hidden representations associated with reflective versus non-reflective reasoning, and find that they exhibit distinct activation patterns. Furthermore, we show that the degree of self-reflection can be modulated by a single direction in the representation space. Our contribution can be summarized as follows.

\begin{itemize} [leftmargin=12pt,itemsep=2pt,topsep=0pt,parsep=0pt]
  \item \textbf{Self-Reflection already emerge during pretraining:} We demonstrate that self-reflection capabilities naturally exist in pretrained models and are not solely artifacts of RLVR. However, the frequency of such behavior is extremely low—for example, only 0.6\% as shown in \Cref{fig:token_distribution}. To isolate the model’s capacity for self-reflection from its general reasoning ability, we propose a method, called {\it \method}, that inserts reasoning traces---specifically, those that trigger self-reflection in a fine-tuned reasoning model---into the input of the pretrained model, and then measures whether the latter produces reflection in response. Using reflection-inducing probing, we observe that the pretrained model exhibits reflection with a frequency of 18.6\%, significantly higher than the baseline of 0.6\%, though still lower than the fine-tuned model (which is almost 100\%).
  Through comparative analysis of hidden representations, we show that pretrained models maintain internal structures that distinguish reflective behavior from non-reflective contexts---similar to fine-tuned models---further suggesting that pretrained models already possess self-reflection capabilities.
  \item \textbf{The degrees of self-reflection can be modulated by a single direction:} Motivated by the separability of reflective and non-reflective contexts in the hidden representation space, we use the method of difference-of-means \cite{rimsky-etal-2024-steering} to construct a {\it self-reflection direction},  enabling control over self-reflection behavior for both pretrained and fine-tuned models.
  Our experiments demonstrate that this control mechanism offers a tunable trade-off between accuracy and efficiency: enhancing reflection improves accuracy by up to 12\% on benchmarks, while suppressing it reduces output length by over 32\% without significant performance degradation. We further show that this direction transfers robustly across diverse tasks—including mathematical and scientific reasoning—highlighting its universality as a shared, task-agnostic mechanism.
\end{itemize}

\section{Preliminary}

\subsection{Transformer Layer}

Decoder-only transformers \cite{liu2018generating} map input tokens $\mathbf{t} = (t_1, t_2, \ldots, t_n) \in \mathcal{V}^n$ to output probability distributions over the vocabulary $\mathcal{V}$. Let $\mathbf{h}_i^{(\ell)} \in \mathbb{R}^d$ denote the residual stream activation (also referred to as the hidden state) of the $i$-th token at the $
\ell$-th layer, where $d$ is the dimensionality of the hidden state. %
Each of the $L$ transformer layers applies a sequence of attention and MLP transformations to update the residual stream:
\e
\tilde{\mathbf{h}}_i^{(\ell)} = \mathbf{h}_i^{(\ell)} + \text{Attn}^{(\ell)}(\mathbf{h}_{1:i}^{(\ell)}), \quad \mathbf{h}_i^{(\ell+1)} = \tilde{\mathbf{h}}_i^{(\ell)} + \text{MLP}^{(\ell)}(\tilde{\mathbf{h}}_i^{(\ell)})
\label{eq:transformer}\ee
The final hidden state is then projected to a probability distribution over the vocabulary $\calV$ using an unembedding matrix followed by a softmax function.

\subsection{Self-Reflection}

Recent work has shown that large language models (LLMs), even when pretrained purely on next-token prediction, demonstrate surprising levels of reasoning ability \cite{mondorf2024beyond, liu2024oats, wang2023selfconsistency}. However, this capability can be significantly enhanced through fine-tuning on reasoning tasks using either reinforcement learning with verifiable rewards (RLVR) or supervised learning with distilled responses from reasoning models trained from RLVR \cite{liu2025x, wang2024reinforcement, zhao2025absolute}. We denote the pretrained model as \pt, and its fine-tuned variant as \ft.

A notable emergent behavior observed in fine-tuned models \ft is self-reflection—the model’s ability to internally evaluate, critique, or revise its own reasoning process. Unlike standard reasoning, which involves generating a direct solution to a task, self-reflection introduces an intermediate meta-cognitive step where the model pauses or backtracks to reconsider its prior outputs. This behavior is often marked by explicit tokens such as “wait,” which have been shown to correlate with improved reasoning outcomes \cite{li2024hindsight, liu2025understanding, yeo2025demystifying}.%

Importantly, self-reflection is not limited to any specific model architecture, observed across both proprietary models \cite{jaech2024openai} and open-source systems \cite{guo2025deepseek, olmo20242, Yang2025Qwen3TR}, indicating that it may be a general emergent property of optimizing for complex reasoning objectives. While models may signal self-reflection using various phrases, including “wait”, “let me double-check”, or “I might have made a mistake”, we focus our analysis on the canonical token “wait” due to its high frequency and clear association with reflective behavior. Our analysis confirms that “wait” is the most commonly used reflection marker across the DeepSeek-R1 series of models
\cite{guo2025deepseek}. Supporting analyses can be found in Appendix~\ref{sec:identify_sr}. Crucially, the analytical framework we develop can generalize beyond "wait", extending to any token that plays an analogous reflective role within the reasoning trajectory.

\section{Self-Reflection Already Emerges During Pretraining}\label{sec:not_unique}

In this section, we conduct a systematic analysis of self-reflection behavior in both pretrained models \ft and fine-tuned ones \pt. Using the MATH500 dataset \cite{hendrycksmath2021} as our evaluation benchmark, we compare DeepSeek-R1-Distill-Qwen-1.5B (\ft), a model fine-tuned from Qwen2.5-1.5B (\pt) \cite{guo2025deepseek, qwen2.5}. First, we show that self-reflection naturally emerges in pretrained models, though at a substantially lower frequency than in RLVR-distilled counterparts. Second, through analysis of hidden state representations, we find that even pretrained models implicitly encode and differentiate between self-reflective and nonself-reflective states, despite rarely generating reflective outputs explicitly.

\begin{figure}
  \includegraphics[width=\linewidth]{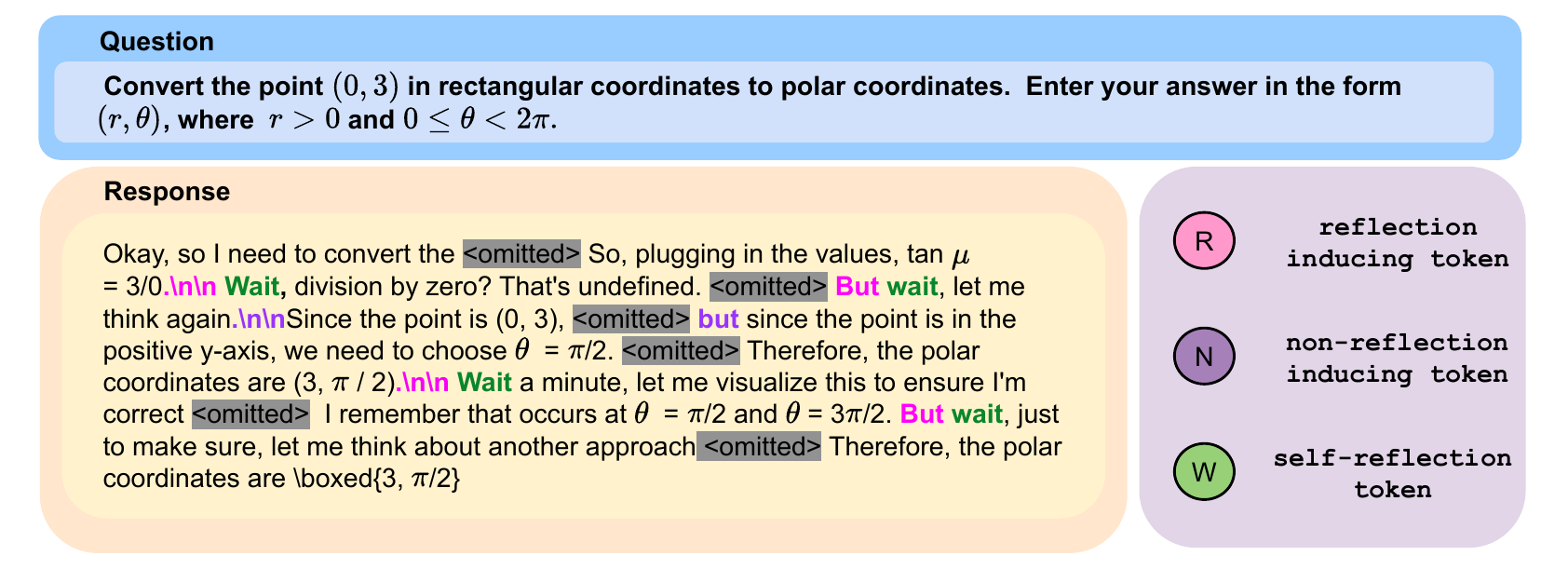}
  \caption{\textbf{Hidden State Selection Methodology:} We identify tokens immediately preceding "wait" tokens as reflection-inducing tokens and extract their hidden states. For comparison, we collect hidden states of identical tokens appearing in non-reflective contexts. This contrastive approach enables us to analyze the neural signatures associated with self-reflection in language models.}
  \label{fig:workflow of hidden state}
\end{figure}

\subsection{Probing Self-Reflection in Pretrained Models}
To investigate whether self-reflection emerges intrinsically in \pt, rather than being solely a byproduct of fine-tuning strategies such as RLVR, we examined the behavior of \pt on mathematical reasoning tasks using the MATH500 benchmark. Figure~\ref{fig:token_distribution} (left) highlights the contrast in self-reflection frequency between \pt and \ft models, while the right panel shows a representative instance of naturally occurring self-reflection in \pt. Remarkably, even in the absence of task-specific supervision, \pt exhibited spontaneous self-reflective behavior in a small but non-negligible fraction of cases—approximately 0.6\%, as shown in Figure~\ref{fig:token_distribution}. These instances are characterized by explicit reconsideration or revision of prior reasoning steps. For details on how we identify self-reflection instances, please see Appendix~\ref{sec:identify_sr}.
While the self-reflection tokens differ somewhat from those typically observed in RLVR-trained models, their reflective nature is still discernible. These findings suggest that self-reflection is not solely acquired through fine-tuning, but rather emerges as a latent capability within the \pt---one that is infrequently activated but nonetheless present.

\paragraph{Reflection-Inducing Probing  by Injecting CoT from \ft into \pt}
However, the extremely low frequency of self-reflection in $A_{\text{pt}}$ makes it challenging to analyze systematically and to develop methods (which will be studied in the next section) for controlling such behavior.
To address this challenge, we propose a probing method, termed {\it \method}, that isolates the model’s capacity for self-reflection from its general reasoning ability. The key idea is to decouple reasoning competence from reflective behavior by inserting reasoning traces generated by the fine-tuned model \ft into the input of the pretrained model \pt, and then measuring whether $\pt$ generates reflection in response.

Formally, given a question $q$, we use the fine-tuned model $\ft$ to generate a sequence of reasoning tokens:
\[
  \ft(q) = (\underbrace{r_1}_{\text{pre-reflection}}, \; \underbrace{\texttt{reflection}}_{\text{explicit signal}}, \; \underbrace{r_2}_{\text{post-reflection}})
\]
where $r_1$ denotes the initial chain-of-thought leading up to an explicit reflection token (e.g., “wait”), and $r_2$ represents the revised or continued reasoning after reflection. We then construct a new prompt by inserting $r_1$ (the pre-reflection reasoning) into the input of \pt, and evaluate whether \pt independently produces a reflection token at the appropriate point. This setup ensures that both models operate on similar reasoning contexts, eliminating confounding differences in reasoning capability. By comparing the frequency and consistency of self-reflection under this controlled setting, we can more directly assess whether reflective behavior is present in the pretrained model and to what extent it is amplified by fine-tuning.
The frequency of generating reflection with \method is reported in Figure~\ref{fig:token_distribution}.

\paragraph{Self-Reflection emerges naturally in pretrained models albeit with much lower frequency}
Remarkably, \pt exhibits clear self-reflective behavior in 18.6\% of these cases—a dramatic increase from its baseline rate. This differential response demonstrates that while \pt rarely produces overt reflection markers in standard contexts, it possesses latent self-reflection capabilities that can be activated by appropriate contextual triggers. These findings strongly suggest that self-reflection mechanisms are encoded during pretraining, rather than being exclusively developed through reinforcement learning. With 18.6\% self-reflection cases, the subsequent section analyzes the hidden state representations underlying these self-reflective behaviors to provide further insights.

\subsection{Hidden State Representations of Self-Reflection}

To further investigate the emergence of self-reflection, we analyze the internal representations of the model when it decides to generate reflection versus when it does not. Specifically, we focus on the hidden states associated with reasoning tokens that immediately precede the generation of a reflection token (e.g., “wait”), and compare them to those that do not lead to reflection.

Since both the pretrained model \pt and the fine-tuned one \ft exhibit self-reflection behaviors, we use \model to denote a generic model (either pretrained or fine-tuned), which will be specified in context. Given a question $q$, suppose the model \model generates a sequence of reasoning tokens that includes self-reflection. Let $r = \model(q) = (r_1, \texttt{reflection}, r_2)$, where $r_1$ precedes the reflection token and $r_2$ follows it. Due to the auto-regressive nature of transformer models, the information from the question $q$ and the reasoning tokens $r_1$ is aggregated into the hidden representation of the final token in $r_1$, which is then used by the last layer to predict the next token---the reflection token. For convenience, we refer to the final token in $r_1$ as a {\it reflection-inducing token}, though its hidden state captures information from the entire preceding context $(q, r_1)$. Reflection-inducing tokens often coincide with sentence-final punctuation (e.g., “.”, “!”, or closing brackets) or specific markers such as "But". Now with a slight abuse of notation, let $h^{(\ell)}_{\text{reflection-inducing}}(q,r)$ denote the $\ell$-th layer hidden state of a model at $\mathcal A$ corresponding to the reflection-inducing token. We collect all such hidden states from model outputs that contain reflection tokens  into the following set
\begin{align}
  \reflect = \left\{ h^{(\ell)}_{\text{reflection-inducing}}(q,r) \in \R^d \right\}.
\end{align}

To study the properties of hidden states associated with self-reflection, we contrast this set with representations from cases where the model {\it does not} generate reflection. Specifically, to eliminate the confounding effect of token surface form, we extract hidden states from tokens that share the same form as reflection-inducing tokens (e.g., sentence-final punctuation), but which do not lead to self-reflection in the subsequent responses (within 100 tokens in the experiments). For notational convenience, we refer to these as {\it non-reflection-inducing tokens}, and denote their corresponding hidden states as $h^{(\ell)}_{\text{non-reflection-inducing}}(q,r)$. See \Cref{fig:workflow of hidden state} for an illustration comparing reflection-inducing and non-reflection-inducing tokens. We collect such non-reflection-inducing into the following set
\begin{align}
  \nonreflect = \left\{ h^{(\ell)}_{\text{non-reflection-inducing}}(q,r) \in \R^d \right\}.
\end{align}
This design ensures a fair comparison by controlling for the surface form of the reflection-inducing token, ensuring that any differences in hidden representations are attributable to the model’s decision to reflect.

\begin{figure}
  \includegraphics[width=\linewidth]{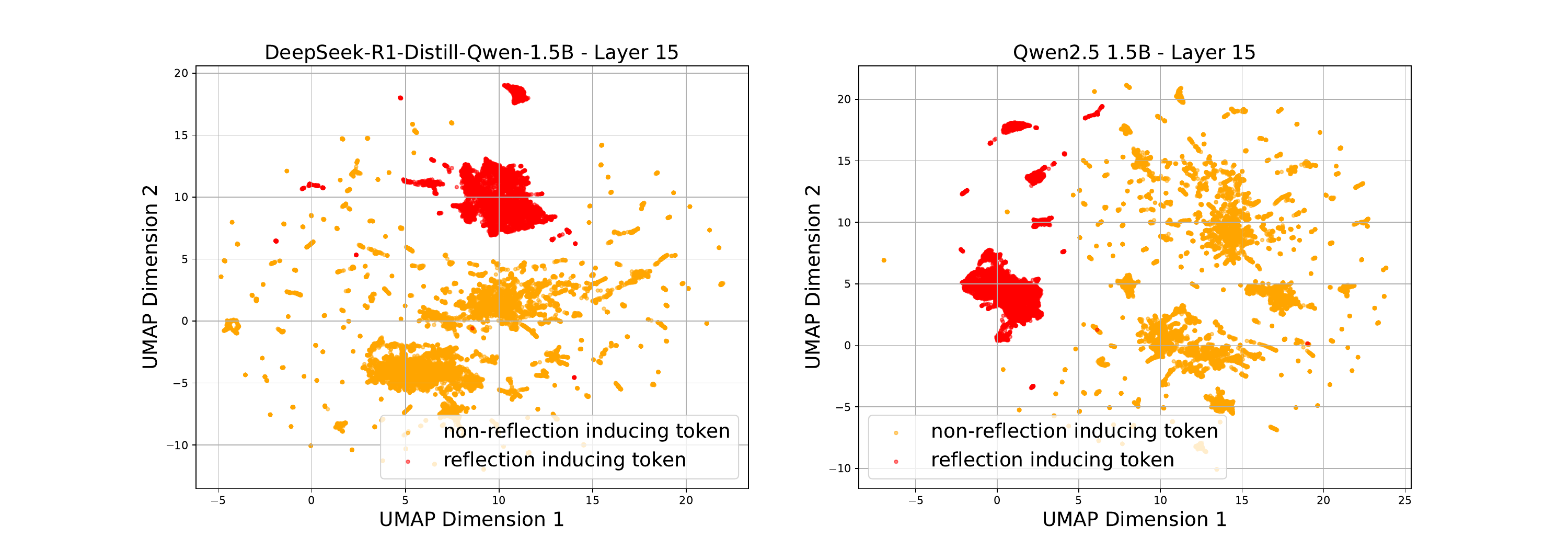}
  \caption{\textbf{UMAP visualization of hidden state representations} of (Left) \ft and (Right) \pt. Both models show separation between \reflect and \nonreflect.
  }
  \label{fig:umap_28}
\end{figure}

For \pt models, which rarely generate self-reflective outputs, we use the method of \method by injecting CoT from \pt into \ft to elicit reflective behavior.
To visualize these high-dimensional representations, we employ UMAP dimensionality reduction \cite{mcinnes2018umap-software}, projecting the hidden states into a 2D space. Figure~\ref{fig:umap_28} presents the visualizations for the 15th layer (out of 28 total layers) for both models, with more layers presented in Appendix~\ref{sec:umap}.

Our analysis reveals a pattern: {\bf both models show clear separation between self-reflection and nonself-reflection states}. While this separation is expected in \ft, which was explicitly trained to exhibit self-reflective behavior, the equally distinct clustering in \pt  is remarkable. Despite rarely generating explicit self-reflection tokens in its outputs, \pt maintains internal representations that clearly distinguish between self-reflective and nonself-reflective contexts. This finding provides strong evidence that self-reflection capabilities develop during pretraining, with models encoding these patterns in their hidden state representations even when they rarely manifest in generated text. We will exploit this internal structure to develop methods for controlling self-reflection in the next section.

\section{Controlling Self-Reflection in Language Models}
In this section, we introduce our approach for identifying and manipulating self-reflection vectors in LLMs. Building on our finding in the last section that hidden representations distinctly separate self-reflective from non-reflective contexts, we construct self-reflection vectors, directions in activation space associated with self-reflective reasoning.
We then demonstrate how these vectors can be used to bidirectionally control self-reflection behavior, either enhancing it to improve reasoning accuracy or suppressing it to reduce computational overhead. Through extensive evaluation across multiple mathematical reasoning benchmarks, we show that our method significantly outperforms strong baselines while offering flexible control over the performance-efficiency trade-off. Finally, we examine the cross-domain transferability of these vectors, revealing their potential as universal controls for self-reflection across diverse reasoning tasks.

\subsection{Extract Self-Reflection Vectors}

To identify the self-reflection vector in the residual stream activations, we compute the difference between the activations of self-reflective and nonself-reflective contexts. This technique, known as difference-in-means, effectively isolates key feature directions, as demonstrated in prior work \cite{rimsky-etal-2024-steering, arditi2024refusal, wu2025axbench}, motivating our application of this approach to the self-reflection domain.
As in Section~\ref{sec:not_unique}, we focus on the hidden state of reflection-inducing token, positing that this state encodes the model's transition into self-reflection reasoning.

For each layer $\ell\in\{1,\dots,L\}$, we compute the mean hidden states in the reflection set \reflect and non-reflection set \nonreflect as
\e
  \refvec
  = \mean(\reflect)
  ,\quad
  \nonrefvec
  = \mean(\nonreflect),
\ee
and then construct the self-reflection vector as
the difference-in-means vector
\begin{align}
  \vv^{(\ell)} = \refvec \;-\; \nonrefvec,
\end{align}
which captures both the direction along which self-reflective and nonself-reflective activations diverge, and the magnitude of that divergence.

\subsection{Model Interventions for Controlling Trade-off between Reasoning and Efficiency}

To actively modulate a model's tendency to reflect, motivated by the linear representation hypothesis and prior work \cite{arditi2024refusal}, we apply simple linear interventions based on the self-reflection vector $\vv^{(\ell)}$ extracted from the $\ell$-th layer, which is expected to capture the direction in representation space most associated with self-reflection. Specifically, we modify each residual stream $\vh^{(\ell)}$ at the $\ell$-th layer in \eqref{eq:transformer} according to
\begin{align}\label{eq:insert}
  \wh\vh^{(\ell)} = \vh^{(\ell)} + \alpha \vv^{(\ell)} \innerprod{\vh^{(\ell)}}{\vv^{(\ell)}},
\end{align}
where $\wh\vh^{(\ell)}$ then replaces $\vh^{(\ell)}$ as the input to the next layer, and the scalar $\alpha$ controls the strength of the intervention. When $\alpha > 0$, the model’s self-reflection behavior is enhanced; when $\alpha < 0$, it is suppressed. Setting $\alpha = 0$ disables the intervention, preserving the model’s default behavior.

\begin{wrapfigure}{r}{0.48\textwidth}
   \centering
   \includegraphics[width=\linewidth]{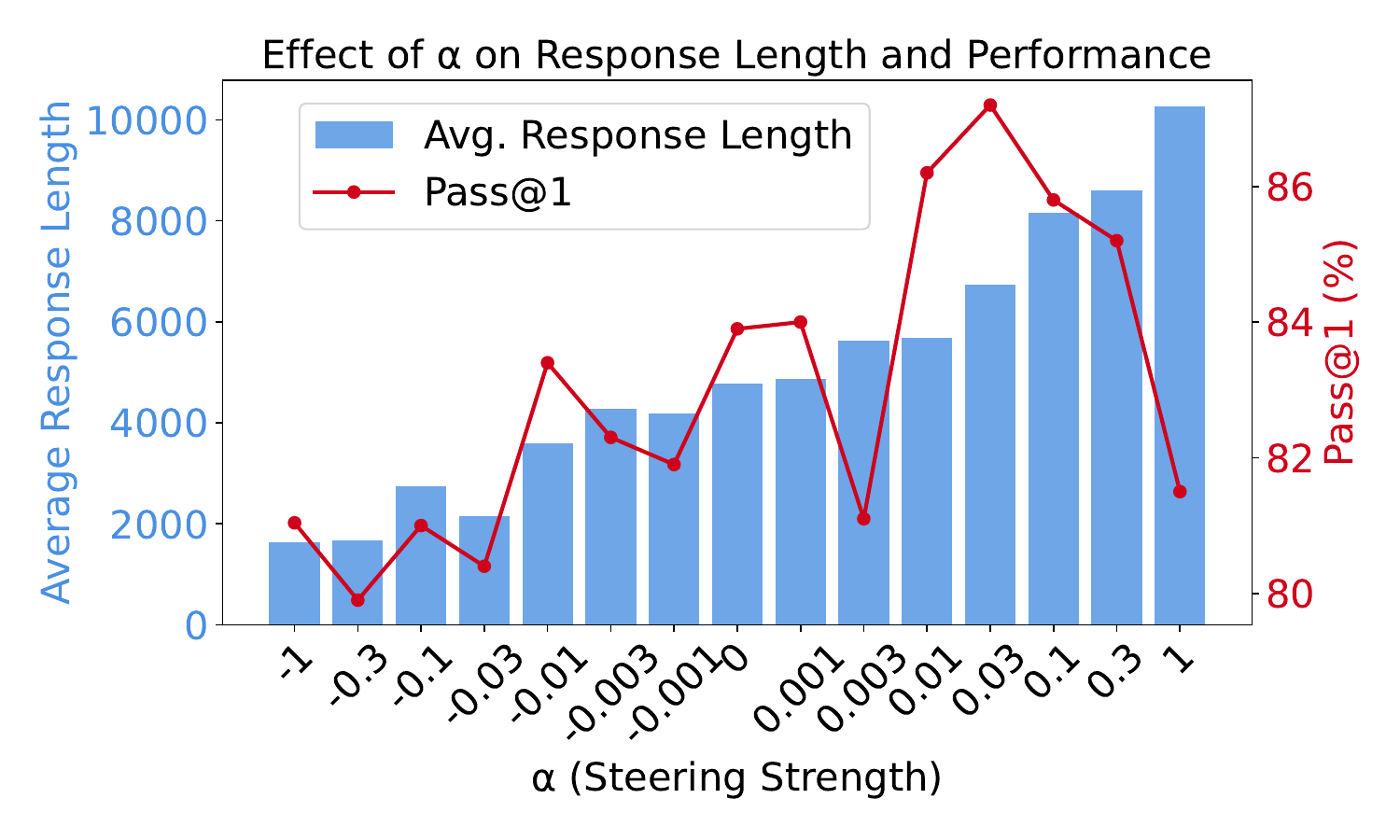}
   \caption{Effect of $\alpha$ on performance and response length on Math500 Dataset (reflection vector is injected on layer 14).}
   \vspace{-.1in}
    \label{fig:alpha_tuning}
\end{wrapfigure}

\paragraph{Ablation study on $\alpha$} To illustrate the effect of the linear intervention method for controlling self-reflection, we conduct an ablation study by varying the self-reflection steering strength $\alpha$ from $-1.0$ to $1.0$, injecting the reflection vector at layer 14 in DeepSeek-R1-1.5B on the MATH-500 benchmark. The result is shown in Figure~\ref{fig:alpha_tuning}. %
Negative $\alpha$ values shorten responses, reducing average token length, while preserving accuracy. In contrast, positive $\alpha$ both lengthens responses, indicating deeper self-reflective reasoning, and boosts performance, peaking at $\alpha$=0.03 with a 12\% performance gain in Pass@1, before declining at larger values due to over‑reflection. This clear trade‑off underscores $\alpha$ as a practical knob for balancing verbosity against reasoning depth. Further ablation detail on the effect of the injection layer is provided in Appendix \ref{app:abl}.

\subsection{Experimental Results}

\begin{table}[t]
  \centering
  \resizebox{\linewidth}{!}{
    \begin{tabular}{clcccccc}
      \toprule
      \multirow{2}{*}{\textbf{Size}} & \multirow{2}{*}{\textbf{Method}} & \multicolumn{2}{c}{\textbf{MATH-500}} & \multicolumn{2}{c}{\textbf{AIME 2024}} & \multicolumn{2}{c}{\textbf{GPQA Diamond}}\\
      \cmidrule(r){3-4} \cmidrule(r){5-6} \cmidrule(r){7-8}
      & & Pass@1 $\uparrow$ & LEN $\downarrow$ & Pass@1 $\uparrow$ & LEN $\downarrow$ & Pass@1 $\uparrow$ & LEN $\downarrow$\\
      \midrule

      \multirow{4}{*}{DeepSeek-R1-1.5B}
      & Vanilla & 84.1 $\pm$ 0.4 & 4755 $\pm$ 216 & 29.2 $\pm$ 1.1 & 6118 $\pm$ 351 & 14.0 $\pm$ 0.9 & 4250 $\pm$ 285 \\
      & BF & 85.5 $\pm$ 0.5 & 10122 $\pm$ 452 & 30.0 $\pm$ 1.3 & 8986 $\pm$ 410 & 14.8 $\pm$ 0.7 & 5210 $\pm$ 339 \\
      & SR Enhanced & \textbf{87.4 $\pm$ 0.3} & 9458 $\pm$ 424 & \textbf{33.5 $\pm$ 1.2} & 8132 $\pm$ 392 & \textbf{18.9 $\pm$ 0.8} & 7496 $\pm$ 462 \\
      & SR Suppressed & 83.2 $\pm$ 0.6 & \textbf{3716 $\pm$ 151} & 27.3 $\pm$ 1.4 & \textbf{5229 $\pm$ 248} & 13.8 $\pm$ 1.0 & \textbf{3795 $\pm$ 199} \\
      \midrule

      \multirow{4}{*}{DeepSeek-R1-7B}
      & Vanilla & 92.7 $\pm$ 0.3 & 3585 $\pm$ 181 & 55.8 $\pm$ 2.0 & 4558 $\pm$ 253 & 27.1 $\pm$ 1.6 & 3696 $\pm$ 215 \\
      & BF & 93.1 $\pm$ 0.2 & 7111 $\pm$ 319 & 54.5 $\pm$ 2.4 & 9629 $\pm$ 488 & 32.0 $\pm$ 1.7 & 5802 $\pm$ 347 \\
      & SR Enhanced & \textbf{93.5 $\pm$ 0.3} & 8959 $\pm$ 402 & \textbf{58.2 $\pm$ 1.9} & 5684 $\pm$ 308 & \textbf{34.6 $\pm$ 1.5} & 6513 $\pm$ 382 \\
      & SR Suppressed & 91.2 $\pm$ 0.4 & \textbf{2439 $\pm$ 118} & 52.9 $\pm$ 2.2 & \textbf{3319 $\pm$ 162} & 26.5 $\pm$ 1.8 & \textbf{3120 $\pm$ 156} \\
      \midrule

      \multirow{3}{*}{Qwen2.5 1.5B}
      & Vanilla & 27.5 $\pm$ 1.5 & \textbf{1515 $\pm$ 98} & 0.2 $\pm$ 0.1 & \textbf{544 $\pm$ 58} & 5.7 $\pm$ 0.8 & \textbf{702 $\pm$ 67} \\
      & BF & 29.6 $\pm$ 1.8 & 3993 $\pm$ 252 & 0.1 $\pm$ 0.1 & 4517 $\pm$ 305 & 5.0 $\pm$ 0.9 & 1389 $\pm$ 114 \\
      & SR Enhanced & \textbf{36.9 $\pm$ 1.4} & 1836 $\pm$ 129 & \textbf{6.1 $\pm$ 0.9} & 2525 $\pm$ 180 & \textbf{6.8 $\pm$ 0.7} & 814 $\pm$ 70 \\
      \midrule

      \multirow{3}{*}{Qwen2.5 7B}
      & Vanilla & 44.8 $\pm$ 2.1 & \textbf{1294 $\pm$ 88} & 3.9 $\pm$ 0.8 & \textbf{2285 $\pm$ 153} & 16.2 $\pm$ 1.4 & \textbf{598 $\pm$ 40} \\
      & BF & 46.0 $\pm$ 1.9 & 3986 $\pm$ 242 & 5.3 $\pm$ 0.9 & 4526 $\pm$ 296 & \textbf{16.9 $\pm$ 1.3} & 1639 $\pm$ 135 \\
      & SR Enhanced & \textbf{56.8 $\pm$ 1.7} & 2671 $\pm$ 174 & \textbf{16.1 $\pm$ 1.4} & 2941 $\pm$ 202 & 14.8 $\pm$ 1.5 & 1816 $\pm$ 147 \\
      \midrule

      \multirow{3}{*}{Llama 3.1 8B Instruct}
      & Vanilla & 44.0 $\pm$ 2.0 & \textbf{1613 $\pm$ 112} & 3.5 $\pm$ 0.6 & \textbf{1653 $\pm$ 125} & 30.9 $\pm$ 1.9 & \textbf{1914 $\pm$ 144} \\
      & BF & 40.1 $\pm$ 2.0 & 3771 $\pm$ 266 & 2.9 $\pm$ 0.7 & 3912 $\pm$ 284 & 26.1 $\pm$ 2.4 & 5031 $\pm$ 351 \\
      & SR Enhanced & \textbf{57.7 $\pm$ 1.6} & 2887 $\pm$ 198 & \textbf{16.5 $\pm$ 1.3} & 2974 $\pm$ 210 & \textbf{34.1 $\pm$ 1.8} & 3995 $\pm$ 310 \\
      \bottomrule
    \end{tabular}
  }
    \caption{Performance across mathematical and scientific reasoning benchmarks using models of different sizes.
  We compare three inference strategies: \textbf{Vanilla} (no intervention), \textbf{BF} (budget forcing via “wait” token insertion), and our method: \textbf{SR Enhanced/Suppressed} (applying positive or negative $\alpha$ to respectively amplify or suppress self-reflection during inference).
  Pass@1 indicates accuracy (higher is better); LEN indicates average generation length (lower is better). We use 10\% of the data as a validation set to select the optimal $\alpha$ in Eq.~\ref{eq:insert}, and report test performance averaged over five runs with different random seeds.
  }
  \label{tab:self-reflect}
\end{table}
We evaluate our self-reflection control mechanism on two mathematical reasoning benchmarks, MATH-500 \cite{hendrycksmath2021} and AIME 2024, and one scientific QA benchmark, GPQA Diamond \cite{rein2024gpqa}. Experiments are conducted using DeepSeek-R1 and Qwen2.5 models at both 1.5B and 7B parameter scales,  as well as the Llama 3.1 8B Instruct model \cite{guo2025deepseek, qwen2.5, grattafiori2024llama}.

We compare three inference strategies:
\textbf{Vanilla (Baseline)}, which uses standard setting without any intervention; \textbf{BF (Budget Forcing)}, which enforces reflection by appending a "wait" token at the end of initial short generations \cite{guo2025deepseek, muennighoff2025s1simpletesttimescaling}; and \textbf{Self-Reflection (SR) Enhanced/Suppressed}, our proposed technique that perturbs hidden states using self-reflection vectors scaled by a coefficient $\alpha$ (positive for enhancement, negative for suppression).
For details on selecting the optimal injection strategy, please refer to Appendix \ref{app:abl}.

\paragraph{Key Findings.}

Our results highlight three major insights. First, SR Enhancement improves reasoning performance across most evaluated datasets and model sizes. For example, Qwen2.5 7B’s performance on MATH‑500 jumps by 12.0 percentage points (from 44.6\% to 56.6\%), and DeepSeek‑R1 variants enjoy similar boosts when employing reflection‑enhanced decoding. Furthermore, this performance gain is also accompanied by a noticeable increase in response length, suggesting that longer responses can be beneficial when tackling more challenging reasoning tasks.

Second, SR Suppression offers fine-grained control over computational cost. It consistently reduces output length—often by more than 50\%—while preserving most of the model’s accuracy. Notably, DeepSeek-R1-7B reduces average token length from 3564 to 2451 on MATH-500 with only a minor drop in Pass@1, which remains above 91\%.

Finally, these effects demonstrate strong generalizability across training paradigms and model families. The observed improvements hold for both \ft models (e.g., DeepSeek-R1) and \pt models (e.g., Qwen2.5, Llama3.1), suggesting that the self-reflection signal is a robust and transferable mechanism that transcends specific architectures or fine-tuning methods.

Building on this generalizability, we emphasize the practical value of inference‑time control over latent self‑reflection dynamics. Unlike rigid interventions such as budget forcing, our method affords semantically grounded, continuous modulation of a model’s internal self-reflection, enabling a tunable trade‑off between performance and efficiency, especially in resource‑constrained settings.

\subsection{Transferability of Self-Reflection Vectors}

\begin{figure}[t]
  \centering
  \begin{minipage}[t]{0.48\textwidth}
    \vspace{0pt}
    \centering
    \includegraphics[width=\linewidth]{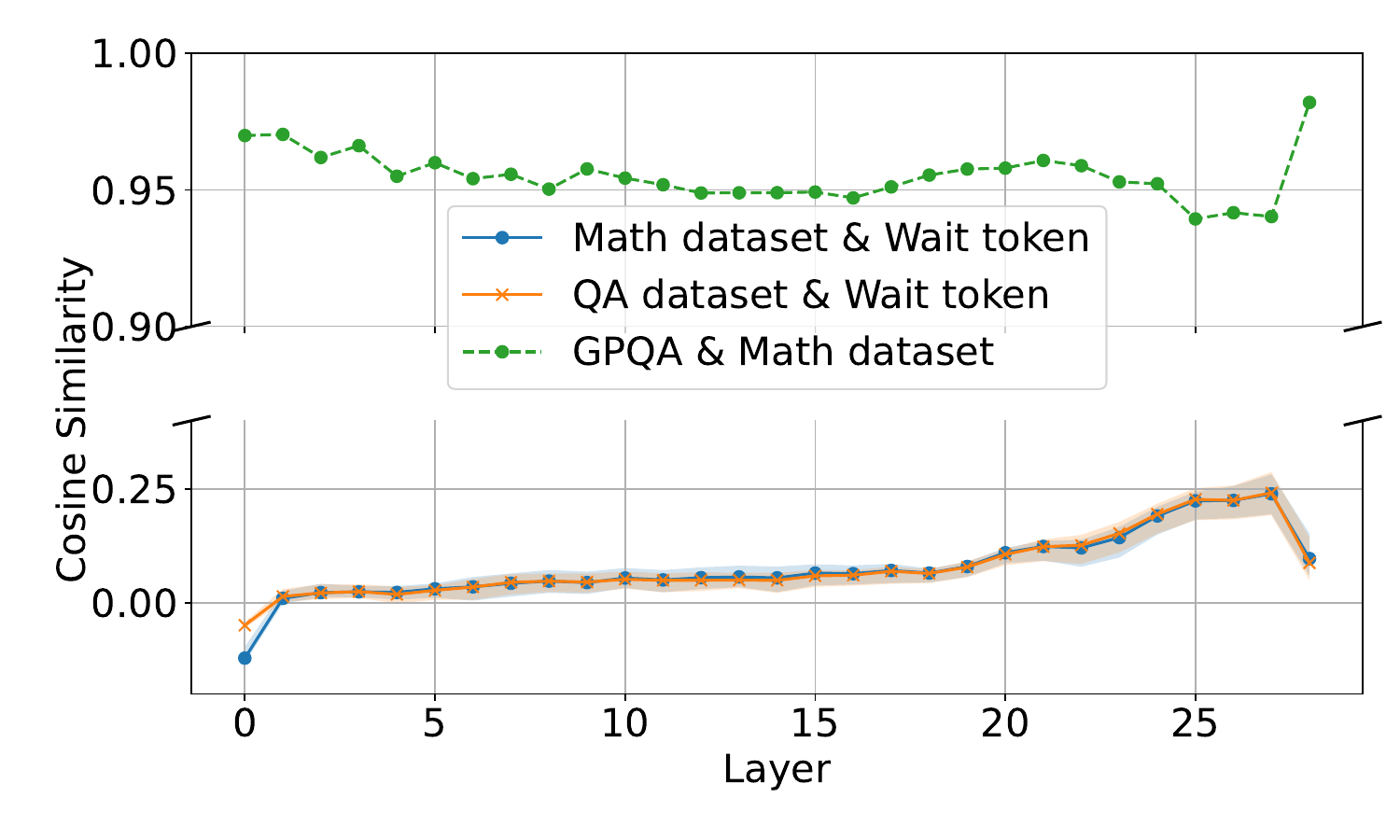}
    \label{fig:cs}
  \end{minipage}%
  \hfill
  \begin{minipage}[t]{0.48\textwidth}
    \vspace{25pt}
    \centering
    \resizebox{\textwidth}{!}{
    \begin{tabular}{llll}
      \hline
      Model & Setting & PASS@1 & Len \\ \hline
      1.5B  & SR Enhanced   & 86.4   & 7781 \\
            & SR Suppressed & 82.6   & 3684 \\
      7B    & SR Enhanced   & 93.0   & 6007 \\
            & SR Suppressed & 91.0   & 2992 \\ \hline
    \end{tabular}}
    \label{tab:transfer}
  \end{minipage}
  \caption{\textbf{Left:} Cosine similarity of self-reflection vectors and the ``wait'' token across MATH500 and GPQA datasets. The green curve shows similarity between vectors from MATH500 and GPQA. Blue and orange curves show similarity with the ``wait'' token. \textbf{Right:} Performance on MATH500 when applying self-reflection vectors extracted from GPQA Diamond to DeepSeek-R1 models.}
\label{fig:transfer}
\end{figure}

To investigate the transferability of self-reflection vectors across different reasoning domains, we evaluated whether vectors extracted from the GPQA Diamond dataset could be effectively transferred to mathematical reasoning tasks in MATH500. We compute the cosine similarity between self-reflection vectors extracted from different domains (MATH500 and GPQA Diamond), and between these vectors and the embedding of the token "wait" in DeepSeek-R1-Distill-Qwen-1.5B. For tokenizers containing multiple subword tokens for "wait", we report the average cosine similarity along with its variance.
The results are plotted in \Cref{fig:transfer}(left).
Our analysis revealed remarkable consistency in the neural signatures of self-reflection across these distinct domains. Specifically, vectors extracted from GPQA and MATH500 exhibit high cosine similarity, suggesting that the internal representation of reflective states is largely domain-invariant. Notably, these self-reflection vectors are substantially different from the embedding of the token "wait", indicating that they encode deeper semantic properties of reflective behavior rather than surface-level cues.

Further, we evaluate the performance of self-reflection vectors derived from GPQA-Diamond on MATH500 using our proposed intervention method, SR-Enhanced/Suppressed, and present the results in \Cref{fig:transfer}(right). Notably, we observe similar performance gains to those seen with in-domain self-reflection vectors, as reported in Table~\ref{tab:self-reflect}.
This cross-domain transfer demonstrates that the reflective mechanism captures a generalizable cognitive pattern rather than being confined to task-specific reasoning strategies. Together, these findings suggest that LLMs develop a unified internal representation of self-reflection, one that can be leveraged across tasks without the need for domain-specific fine-tuning.

\section{Related Work}

\subsection{Features as Directions}
Extracting feature directions, often derived from contrastive pairs of inputs, is an established technique for analyzing and manipulating neural network representations \cite{rimsky-etal-2024-steering, burns2022dl, zou2023universal}.
It is widely recognized that adding such feature vectors to the model's residual stream can modify its behavior, although the optimal intervention points and specific methodologies remain areas of active research \cite{von2024language, jorgensen2023improving}.

Several studies suggest that directions within the activation space capture semantic features more effectively or interpretably than individual neurons \cite{geiger2024finding, Park2023GenerativeAgents, bolukbasi2016man}.
Recent approaches utilize techniques like sparse autoencoders to discover these feature directions in an supervised manner \cite{huben2024sparse}.
However, alternative methods such as Difference-in-Means (DiffMean) \cite{arditi2024refusal} have demonstrated strong performance, sometimes exceeding that of sparse autoencoders, in specific applications like concept detection and model steering \cite{wu2025axbench}.
Furthermore, the underlying assumption that features can be represented linearly has proven effective in tasks such as targeted concept erasure within language models \cite{shao-etal-2023-gold, belrose2023leace, feng2025monitoring}.

\subsection{Self-Reflection in Language Models}
The concept of self-reflection in language models has gained increasing attention as a mechanism for improving reasoning quality and alignment. Recent studies \cite{lightman2024lets, madaan2023selfrefine, puerto2024finetuning, zelikman2022star, lightman2023let, li2025llms} have explored how prompting models to generate intermediate reflections, critiques, or alternative solutions can improve final outputs in tasks such as math problem solving, programming, and factual reasoning. While many of these techniques are implemented at the prompting level or through chain-of-thought scaffolding, they suggest that self-reflection is a powerful tool for enhancing reasoning. Notably, these methods often induce substantial increases in generation length and latency, raising questions about the trade-off between deliberation and efficiency \cite{yang2025dynamic, yi2025shorterbetter, chen2024unlocking, team2025kimi}.

Recent studies have show that RLVR \cite{shah2025rethinking, shinn2023reflexion, wang2025reinforcement, xu2025towards} can improve reasoning abilities by explicitly training models to reflect using outcome-based feedback.
Empirically, trained models \cite{guo2025deepseek, liu2025there} like DeepSeek-R1 demonstrate significant improvements over baseline models in mathematical and logical reasoning tasks, and showcase new emergent behaviors such as self-reflection. Our work shows that self-reflection is a broadly distributed and latent feature of LLMs, not exclusively a product of RLVR. Concurrent works \cite{ shah2025rethinking, yue2025does} also suggest that RLVR does not necessarily introduce novel reasoning abilities beyond those acquired during pretraining; instead, it primarily serves to amplify abilities already present in the model. Our work also complements this literature by showing that LLMs already encode latent self-reflection signals in their hidden states—even in models not explicitly trained for such behavior—and that reflection can be selectively enhanced or suppressed through lightweight vector interventions. This enables fine-grained control over reflective behavior, including the ability to mitigate over-reflection, thereby avoiding unnecessary computational overhead without sacrificing performance.

\section{Conclusion}

In this paper, we demonstrated that self-reflection in large language models is an emergent capability that develops during pretraining rather than being uniquely induced by reinforcement learning techniques. Through contrastive analysis of hidden state representations, we revealed that even models with minimal explicit reflection behavior maintain internal neural signatures that distinguish self-reflective contexts. By exploiting these representations, we developed an intervention method that enables bidirectional control over self-reflection, providing a flexible mechanism to navigate the performance-efficiency trade-off without requiring additional training.

\section*{Acknowledgement}
We acknowledge support from NSF grants IIS-2312840 and IIS-2402952, as well
as the ORAU Ralph E. Powe Junior Faculty Enhancement Award. The authors thank Vishnu Chhabra and Suchit Gupte for valuable discussions.

\bibliography{ref}

\begin{thebibliography}{10}

\bibitem{xu2025towards}
F.~Xu, Q.~Hao, Z.~Zong, J.~Wang, Y.~Zhang, J.~Wang, X.~Lan, J.~Gong, T.~Ouyang, F.~Meng, {\em et~al.}, ``Towards large reasoning models: A survey of reinforced reasoning with large language models,'' {\em arXiv preprint arXiv:2501.09686}, 2025.

\bibitem{wang2025selfreasoning}
H.~WANG, D.~Cai, W.~Zhong, S.~Huang, J.~Z. Pan, Z.~Liu, and K.-F. Wong, ``Self-reasoning language models: Unfold hidden reasoning chains with few reasoning catalyst,'' in {\em Workshop on Reasoning and Planning for Large Language Models}, 2025.

\bibitem{mroueh2025reinforcement}
Y.~Mroueh, ``Reinforcement learning with verifiable rewards: Grpo's effective loss, dynamics, and success amplification,'' {\em arXiv preprint arXiv:2503.06639}, 2025.

\bibitem{zhao2025r1}
J.~Zhao, X.~Wei, and L.~Bo, ``R1-omni: Explainable omni-multimodal emotion recognition with reinforcement learning,'' {\em arXiv preprint arXiv:2503.05379}, 2025.

\bibitem{ferrag2025reasoning}
M.~A. Ferrag, N.~Tihanyi, and M.~Debbah, ``Reasoning beyond limits: Advances and open problems for llms,'' {\em arXiv preprint arXiv:2503.22732}, 2025.

\bibitem{su2025expanding}
Y.~Su, D.~Yu, L.~Song, J.~Li, H.~Mi, Z.~Tu, M.~Zhang, and D.~Yu, ``Expanding rl with verifiable rewards across diverse domains,'' {\em arXiv preprint arXiv:2503.23829}, 2025.

\bibitem{guo2025deepseek}
D.~Guo, D.~Yang, H.~Zhang, J.~Song, R.~Zhang, R.~Xu, Q.~Zhu, S.~Ma, P.~Wang, X.~Bi, {\em et~al.}, ``Deepseek-r1: Incentivizing reasoning capability in llms via reinforcement learning,'' {\em arXiv preprint arXiv:2501.12948}, 2025.

\bibitem{liu2025understanding}
Z.~Liu, C.~Chen, W.~Li, P.~Qi, T.~Pang, C.~Du, W.~S. Lee, and M.~Lin, ``Understanding r1-zero-like training: A critical perspective,'' {\em arXiv preprint arXiv:2503.20783}, 2025.

\bibitem{zeng2025simplerl}
W.~Zeng, Y.~Huang, Q.~Liu, W.~Liu, K.~He, Z.~Ma, and J.~He, ``Simplerl-zoo: Investigating and taming zero reinforcement learning for open base models in the wild,'' {\em arXiv preprint arXiv:2503.18892}, 2025.

\bibitem{zuo2025ttrl}
Y.~Zuo, K.~Zhang, S.~Qu, L.~Sheng, X.~Zhu, B.~Qi, Y.~Sun, G.~Cui, N.~Ding, and B.~Zhou, ``Ttrl: Test-time reinforcement learning,'' {\em arXiv preprint arXiv:2504.16084}, 2025.

\bibitem{yue2025does}
Y.~Yue, Z.~Chen, R.~Lu, A.~Zhao, Z.~Wang, S.~Song, and G.~Huang, ``Does reinforcement learning really incentivize reasoning capacity in llms beyond the base model?,'' {\em arXiv preprint arXiv:2504.13837}, 2025.

\bibitem{liu2024self}
F.~Liu, N.~AlDahoul, G.~Eady, Y.~Zaki, B.~AlShebli, and T.~Rahwan, ``Self-reflection outcome is sensitive to prompt construction,'' {\em arXiv preprint arXiv:2406.10400}, 2024.

\bibitem{yang2025dynamic}
C.~Yang, Q.~Si, Y.~Duan, Z.~Zhu, C.~Zhu, Z.~Lin, L.~Cao, and W.~Wang, ``Dynamic early exit in reasoning models,'' {\em arXiv preprint arXiv:2504.15895}, 2025.

\bibitem{renze2024self}
M.~Renze and E.~Guven, ``Self-reflection in llm agents: Effects on problem-solving performance,'' {\em arXiv preprint arXiv:2405.06682}, 2024.

\bibitem{sui2025stop}
Y.~Sui, Y.-N. Chuang, G.~Wang, J.~Zhang, T.~Zhang, J.~Yuan, H.~Liu, A.~Wen, S.~Zhong, H.~Chen, {\em et~al.}, ``Stop overthinking: A survey on efficient reasoning for large language models,'' {\em arXiv preprint arXiv:2503.16419}, 2025.

\bibitem{rimsky-etal-2024-steering}
N.~Rimsky, N.~Gabrieli, J.~Schulz, M.~Tong, E.~Hubinger, and A.~Turner, ``Steering llama 2 via contrastive activation addition,'' in {\em Proceedings of the 62nd Annual Meeting of the Association for Computational Linguistics (Volume 1: Long Papers)} (L.-W. Ku, A.~Martins, and V.~Srikumar, eds.), (Bangkok, Thailand), pp.~15504--15522, Association for Computational Linguistics, Aug. 2024.

\bibitem{liu2018generating}
P.~J. Liu, M.~Saleh, E.~Pot, B.~Goodrich, R.~Sepassi, L.~Kaiser, and N.~Shazeer, ``Generating wikipedia by summarizing long sequences,'' {\em arXiv preprint arXiv:1801.10198}, 2018.

\bibitem{mondorf2024beyond}
P.~Mondorf and B.~Plank, ``Beyond accuracy: Evaluating the reasoning behavior of large language models - a survey,'' in {\em First Conference on Language Modeling}, 2024.

\bibitem{liu2024oats}
Z.~Liu, C.~Chen, C.~Du, W.~S. Lee, and M.~Lin, ``Oat: A research-friendly framework for llm online alignment,'' 2024.

\bibitem{wang2023selfconsistency}
X.~Wang, J.~Wei, D.~Schuurmans, Q.~V. Le, E.~H. Chi, S.~Narang, A.~Chowdhery, and D.~Zhou, ``Self-consistency improves chain of thought reasoning in language models,'' in {\em The Eleventh International Conference on Learning Representations}, 2023.

\bibitem{liu2025x}
Q.~Liu, S.~Zhang, G.~Qin, T.~Ossowski, Y.~Gu, Y.~Jin, S.~Kiblawi, S.~Preston, M.~Wei, P.~Vozila, T.~Naumann, and H.~Poon, ``X-reasoner: Towards generalizable reasoning across modalities and domains,'' 2025.

\bibitem{wang2024reinforcement}
S.~Wang, S.~Zhang, J.~Zhang, R.~Hu, X.~Li, T.~Zhang, J.~Li, F.~Wu, G.~Wang, and E.~Hovy, ``Reinforcement learning enhanced llms: A survey,'' {\em arXiv preprint arXiv:2412.10400}, 2024.

\bibitem{zhao2025absolute}
A.~Zhao, Y.~Wu, Y.~Yue, T.~Wu, Q.~Xu, M.~Lin, S.~Wang, Q.~Wu, Z.~Zheng, and G.~Huang, ``Absolute zero: Reinforced self-play reasoning with zero data,'' {\em arXiv preprint arXiv:2505.03335}, 2025.

\bibitem{li2024hindsight}
Y.~Li, C.~Yang, and A.~Ettinger, ``When hindsight is not 20/20: Testing limits on reflective thinking in large language models,'' {\em arXiv preprint arXiv:2404.09129}, 2024.

\bibitem{yeo2025demystifying}
E.~Yeo, Y.~Tong, X.~Niu, G.~Neubig, and X.~Yue, ``Demystifying long chain-of-thought reasoning in {LLM}s,'' in {\em ICLR 2025 Workshop on Navigating and Addressing Data Problems for Foundation Models}, 2025.

\bibitem{jaech2024openai}
A.~Jaech, A.~Kalai, A.~Lerer, A.~Richardson, A.~El-Kishky, A.~Low, A.~Helyar, A.~Madry, A.~Beutel, A.~Carney, {\em et~al.}, ``Openai o1 system card,'' {\em arXiv preprint arXiv:2412.16720}, 2024.

\bibitem{olmo20242}
T.~OLMo, P.~Walsh, L.~Soldaini, D.~Groeneveld, K.~Lo, S.~Arora, A.~Bhagia, Y.~Gu, S.~Huang, M.~Jordan, {\em et~al.}, ``2 olmo 2 furious,'' {\em arXiv preprint arXiv:2501.00656}, 2024.

\bibitem{Yang2025Qwen3TR}
A.~Yang, A.~Li, B.~Yang, B.~Zhang, B.~Hui, B.~Zheng, B.~Yu, C.~Gao, C.~Huang, C.~Lv, C.~Zheng, D.~Liu, F.~Zhou, F.~Huang, F.~Hu, H.~Ge, H.~Wei, H.~Lin, J.~Tang, J.~Yang, J.~Tu, J.~Zhang, J.~Yang, J.~Yang, J.~Zhou, J.~Zhou, J.~Lin, K.~Dang, K.~Bao, K.~Yang, L.~Yu, L.~Deng, M.~Li, M.~Xue, M.~Li, P.~Zhang, P.~Wang, Q.~Zhu, R.~Men, R.~Gao, S.~Liu, S.~Luo, T.~Li, T.~Tang, W.~Yin, X.~Ren, X.~Wang, X.~Zhang, X.~Ren, Y.~Fan, Y.~Su, Y.-C. Zhang, Y.~Zhang, Y.~Wan, Y.~Liu, Z.~Wang, Z.~Cui, Z.~Zhang, Z.~Zhou, and Z.~Qiu, ``Qwen3 technical report,'' 2025.

\bibitem{hendrycksmath2021}
D.~Hendrycks, C.~Burns, S.~Kadavath, A.~Arora, S.~Basart, E.~Tang, D.~Song, and J.~Steinhardt, ``Measuring mathematical problem solving with the math dataset,'' {\em NeurIPS}, 2021.

\bibitem{qwen2.5}
A.~Yang, B.~Yang, B.~Zhang, B.~Hui, B.~Zheng, B.~Yu, C.~Li, D.~Liu, F.~Huang, H.~Wei, H.~Lin, J.~Yang, J.~Tu, J.~Zhang, J.~Yang, J.~Yang, J.~Zhou, J.~Lin, K.~Dang, K.~Lu, K.~Bao, K.~Yang, L.~Yu, M.~Li, M.~Xue, P.~Zhang, Q.~Zhu, R.~Men, R.~Lin, T.~Li, T.~Xia, X.~Ren, X.~Ren, Y.~Fan, Y.~Su, Y.~Zhang, Y.~Wan, Y.~Liu, Z.~Cui, Z.~Zhang, and Z.~Qiu, ``Qwen2.5 technical report,'' {\em arXiv preprint arXiv:2412.15115}, 2024.

\bibitem{mcinnes2018umap-software}
L.~McInnes, J.~Healy, N.~Saul, and L.~Grossberger, ``Umap: Uniform manifold approximation and projection,'' {\em The Journal of Open Source Software}, vol.~3, no.~29, p.~861, 2018.

\bibitem{arditi2024refusal}
A.~Arditi, O.~B. Obeso, A.~Syed, D.~Paleka, N.~Rimsky, W.~Gurnee, and N.~Nanda, ``Refusal in language models is mediated by a single direction,'' in {\em The Thirty-eighth Annual Conference on Neural Information Processing Systems}, 2024.

\bibitem{wu2025axbench}
Z.~Wu, A.~Arora, A.~Geiger, Z.~Wang, J.~Huang, D.~Jurafsky, C.~D. Manning, and C.~Potts, ``Axbench: Steering llms? even simple baselines outperform sparse autoencoders,'' {\em arXiv preprint arXiv:2501.17148}, 2025.

\bibitem{rein2024gpqa}
D.~Rein, B.~L. Hou, A.~C. Stickland, J.~Petty, R.~Y. Pang, J.~Dirani, J.~Michael, and S.~R. Bowman, ``{GPQA}: A graduate-level google-proof q\&a benchmark,'' in {\em First Conference on Language Modeling}, 2024.

\bibitem{grattafiori2024llama}
A.~Grattafiori, A.~Dubey, A.~Jauhri, A.~Pandey, A.~Kadian, A.~Al-Dahle, A.~Letman, A.~Mathur, A.~Schelten, A.~Vaughan, {\em et~al.}, ``The llama 3 herd of models,'' {\em arXiv preprint arXiv:2407.21783}, 2024.

\bibitem{muennighoff2025s1simpletesttimescaling}
N.~Muennighoff, Z.~Yang, W.~Shi, X.~L. Li, L.~Fei-Fei, H.~Hajishirzi, L.~Zettlemoyer, P.~Liang, E.~Candès, and T.~Hashimoto, ``s1: Simple test-time scaling,'' 2025.

\bibitem{burns2022dl}
C.~Burns, H.~Ye, D.~Klein, and J.~Steinhardt, ``Discovering latent knowledge in language models without supervision,'' {\em ArXiV}, 2022.

\bibitem{zou2023universal}
A.~Zou, Z.~Wang, N.~Carlini, M.~Nasr, J.~Z. Kolter, and M.~Fredrikson, ``Universal and transferable adversarial attacks on aligned language models,'' {\em arXiv preprint arXiv:2307.15043}, 2023.

\bibitem{von2024language}
D.~Von~R{\"u}tte, S.~Anagnostidis, G.~Bachmann, and T.~Hofmann, ``A language model's guide through latent space,'' {\em arXiv preprint arXiv:2402.14433}, 2024.

\bibitem{jorgensen2023improving}
O.~Jorgensen, D.~Cope, N.~Schoots, and M.~Shanahan, ``Improving activation steering in language models with mean-centring,'' {\em arXiv preprint arXiv:2312.03813}, 2023.

\bibitem{geiger2024finding}
A.~Geiger, Z.~Wu, C.~Potts, T.~Icard, and N.~Goodman, ``Finding alignments between interpretable causal variables and distributed neural representations,'' in {\em Causal Learning and Reasoning}, pp.~160--187, PMLR, 2024.

\bibitem{Park2023GenerativeAgents}
J.~S. Park, J.~C. O'Brien, C.~J. Cai, M.~R. Morris, P.~Liang, and M.~S. Bernstein, ``Generative agents: Interactive simulacra of human behavior,'' in {\em In the 36th Annual ACM Symposium on User Interface Software and Technology (UIST '23)}, UIST '23, (New York, NY, USA), Association for Computing Machinery, 2023.

\bibitem{bolukbasi2016man}
T.~Bolukbasi, K.-W. Chang, J.~Y. Zou, V.~Saligrama, and A.~T. Kalai, ``Man is to computer programmer as woman is to homemaker? debiasing word embeddings,'' {\em Advances in neural information processing systems}, vol.~29, 2016.

\bibitem{huben2024sparse}
R.~Huben, H.~Cunningham, L.~R. Smith, A.~Ewart, and L.~Sharkey, ``Sparse autoencoders find highly interpretable features in language models,'' in {\em The Twelfth International Conference on Learning Representations}, 2024.

\bibitem{shao-etal-2023-gold}
S.~Shao, Y.~Ziser, and S.~B. Cohen, ``Gold doesn`t always glitter: Spectral removal of linear and nonlinear guarded attribute information,'' in {\em Proceedings of the 17th Conference of the European Chapter of the Association for Computational Linguistics} (A.~Vlachos and I.~Augenstein, eds.), (Dubrovnik, Croatia), pp.~1611--1622, Association for Computational Linguistics, May 2023.

\bibitem{belrose2023leace}
N.~Belrose, D.~Schneider-Joseph, S.~Ravfogel, R.~Cotterell, E.~Raff, and S.~Biderman, ``{LEACE}: Perfect linear concept erasure in closed form,'' in {\em Thirty-seventh Conference on Neural Information Processing Systems}, 2023.

\bibitem{feng2025monitoring}
J.~Feng, S.~Russell, and J.~Steinhardt, ``Monitoring latent world states in language models with propositional probes,'' in {\em The Thirteenth International Conference on Learning Representations}, 2025.

\bibitem{lightman2024lets}
H.~Lightman, V.~Kosaraju, Y.~Burda, H.~Edwards, B.~Baker, T.~Lee, J.~Leike, J.~Schulman, I.~Sutskever, and K.~Cobbe, ``Let's verify step by step,'' in {\em The Twelfth International Conference on Learning Representations}, 2024.

\bibitem{madaan2023selfrefine}
A.~Madaan, N.~Tandon, P.~Gupta, S.~Hallinan, L.~Gao, S.~Wiegreffe, U.~Alon, N.~Dziri, S.~Prabhumoye, Y.~Yang, S.~Gupta, B.~P. Majumder, K.~Hermann, S.~Welleck, A.~Yazdanbakhsh, and P.~Clark, ``Self-refine: Iterative refinement with self-feedback,'' in {\em Thirty-seventh Conference on Neural Information Processing Systems}, 2023.

\bibitem{puerto2024finetuning}
H.~Puerto, T.~Chubakov, X.~Zhu, H.~T. Madabushi, and I.~Gurevych, ``Fine-tuning with divergent chains of thought boosts reasoning through self-correction in language models,'' 2024.

\bibitem{zelikman2022star}
E.~Zelikman, Y.~Wu, J.~Mu, and N.~Goodman, ``{ST}ar: Bootstrapping reasoning with reasoning,'' in {\em Advances in Neural Information Processing Systems} (A.~H. Oh, A.~Agarwal, D.~Belgrave, and K.~Cho, eds.), 2022.

\bibitem{lightman2023let}
H.~Lightman, V.~Kosaraju, Y.~Burda, H.~Edwards, B.~Baker, T.~Lee, J.~Leike, J.~Schulman, I.~Sutskever, and K.~Cobbe, ``Let's verify step by step,'' in {\em The Twelfth International Conference on Learning Representations}, 2023.

\bibitem{li2025llms}
D.~Li, S.~Cao, T.~Griggs, S.~Liu, X.~Mo, E.~Tang, S.~Hegde, K.~Hakhamaneshi, S.~G. Patil, M.~Zaharia, {\em et~al.}, ``Llms can easily learn to reason from demonstrations structure, not content, is what matters!,'' {\em arXiv preprint arXiv:2502.07374}, 2025.

\bibitem{yi2025shorterbetter}
J.~Yi and J.~Wang, ``Shorterbetter: Guiding reasoning models to find optimal inference length for efficient reasoning,'' {\em arXiv preprint arXiv:2504.21370}, 2025.

\bibitem{chen2024unlocking}
Q.~Chen, L.~Qin, J.~WANG, J.~Zhou, and W.~Che, ``Unlocking the capabilities of thought: A reasoning boundary framework to quantify and optimize chain-of-thought,'' in {\em The Thirty-eighth Annual Conference on Neural Information Processing Systems}, 2024.

\bibitem{team2025kimi}
K.~Team, A.~Du, B.~Gao, B.~Xing, C.~Jiang, C.~Chen, C.~Li, C.~Xiao, C.~Du, C.~Liao, {\em et~al.}, ``Kimi k1. 5: Scaling reinforcement learning with llms,'' {\em arXiv preprint arXiv:2501.12599}, 2025.

\bibitem{shah2025rethinking}
D.~J. Shah, P.~Rushton, S.~Singla, M.~Parmar, K.~Smith, Y.~Vanjani, A.~Vaswani, A.~Chaluvaraju, A.~Hojel, A.~Ma, {\em et~al.}, ``Rethinking reflection in pre-training,'' {\em arXiv preprint arXiv:2504.04022}, 2025.

\bibitem{shinn2023reflexion}
N.~Shinn, F.~Cassano, A.~Gopinath, K.~Narasimhan, and S.~Yao, ``Reflexion: Language agents with verbal reinforcement learning,'' {\em Advances in Neural Information Processing Systems}, vol.~36, pp.~8634--8652, 2023.

\bibitem{wang2025reinforcement}
Y.~Wang, Q.~Yang, Z.~Zeng, L.~Ren, L.~Liu, B.~Peng, H.~Cheng, X.~He, K.~Wang, J.~Gao, {\em et~al.}, ``Reinforcement learning for reasoning in large language models with one training example,'' {\em arXiv preprint arXiv:2504.20571}, 2025.

\bibitem{liu2025there}
Z.~Liu, C.~Chen, W.~Li, T.~Pang, C.~Du, and M.~Lin, ``There may not be aha moment in r1-zero-like training — a pilot study.'' \url{https://oatllm.notion.site/oat-zero}, 2025.
\newblock Notion Blog.

\end{thebibliography}
\bibliographystyle{ieeetr}

\newpage

\appendix

\section{UMAP Visualization of Self-Reflection States}
\label{sec:umap}

\begin{figure}[!htp]
  \centering
  \includegraphics[width=0.8\textwidth]{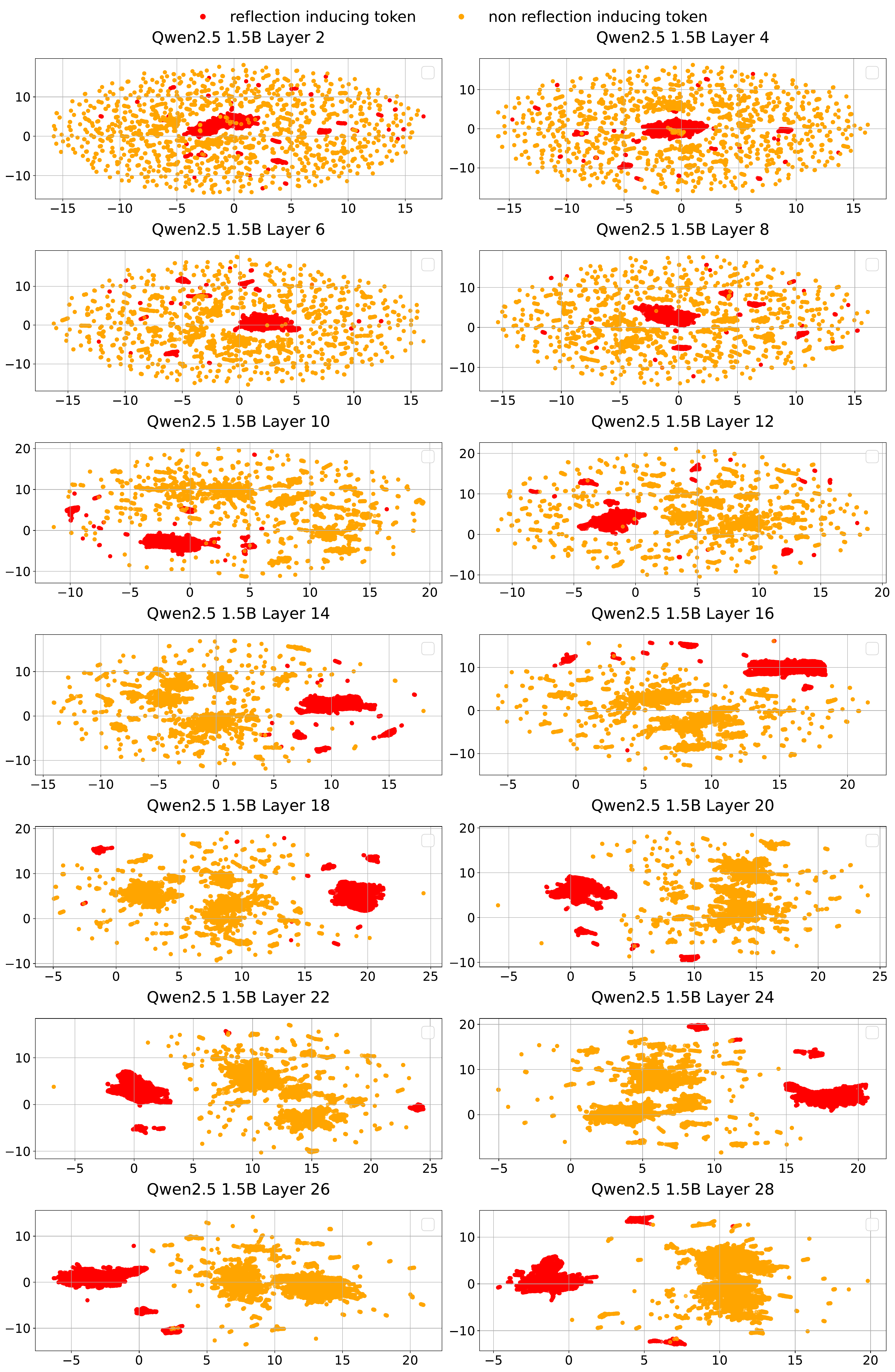}
  \caption{UMAP visualization of hidden states from the Qwen2.5-1.5B model across 14 layers.}
  \label{fig:all_layers_umap_qwen}
\end{figure}

\begin{figure}[!htp]
  \centering
  \includegraphics[width=0.8\textwidth]{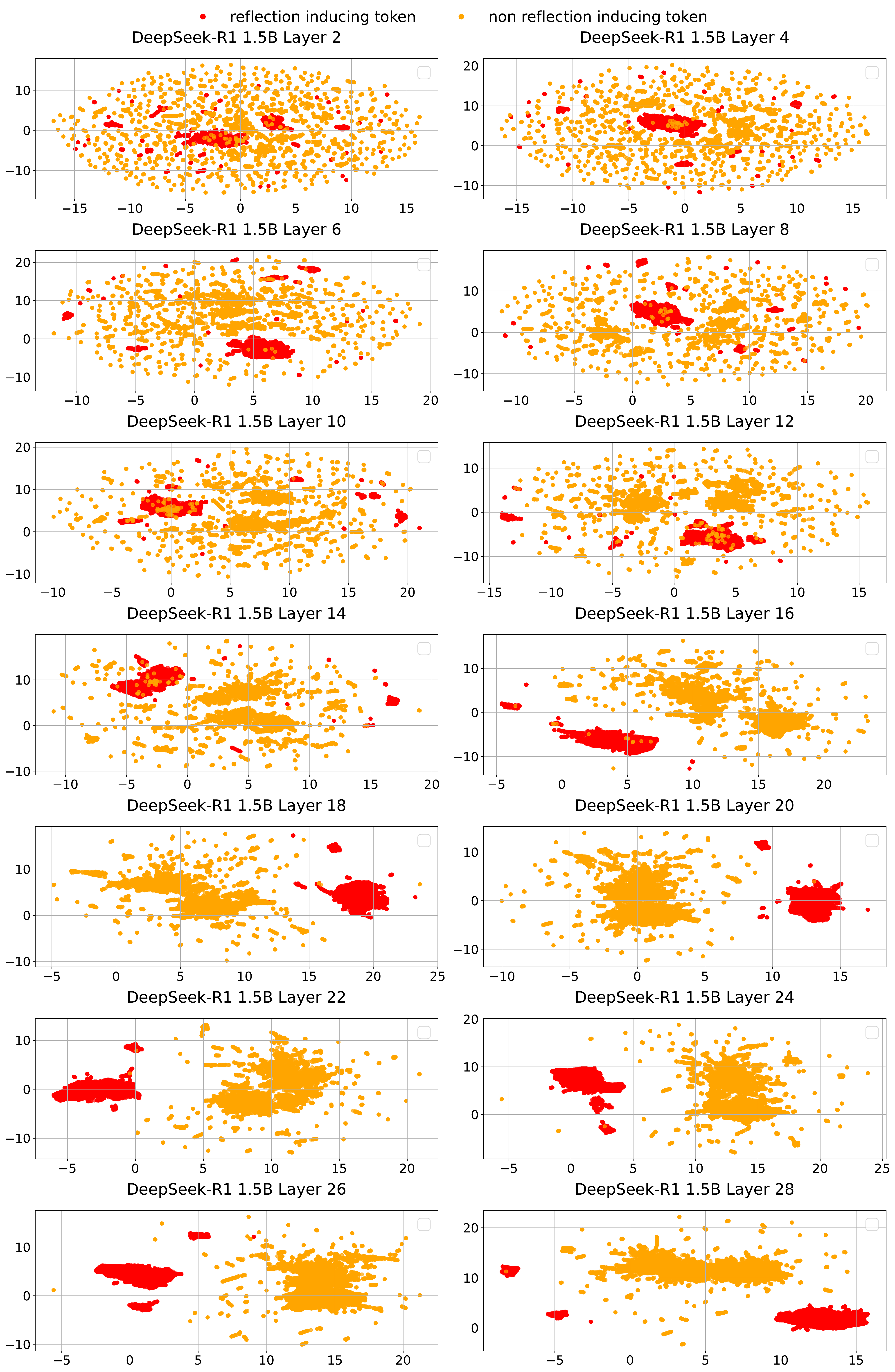}
  \caption{UMAP visualization of hidden states from the DeepSeek-R1-Distill-Qwen-1.5B model across 14 layers.}
  \label{fig:all_layers_umap_ds}
\end{figure}

To investigate how self-reflective states are internally represented, we perform UMAP-based dimensionality reduction on hidden states extracted from both Qwen2.5 1.5B and DeepSeek-R1-Distill-Qwen-1.5B models. Figure~\ref{fig:umap_28} in the main text illustrates the final-layer representations, while Figures~\ref{fig:all_layers_umap_qwen} and~\ref{fig:all_layers_umap_ds} offer a representative view of 14 out of 28 layers across the Qwen2.5-1.5B and DeepSeek-R1-Distill-Qwen-1.5B models, respectively.

We observe that the separation between reflective and nonself-reflective hidden states is not binary, but rather continuous. As we move from shallow to deeper layers, the distinction becomes more pronounced. Early layers show considerable overlap between reflective and non-reflective tokens. In middle layers, clustering begins to emerge, and by the final layers, the separation becomes clear and robust. This progression implies that self-reflective encoding is a hierarchical feature—gradually constructed through successive transformations, rather than being localized to a single depth.

Notably, even in Qwen2.5, which rarely emits explicit self-reflection tokens, the hidden states exhibit consistent separability. This finding reinforces our central hypothesis: the capacity for self-reflection is not merely a byproduct of reinforcement learning or specialized fine-tuning (e.g., RLVR), but an inherent representational feature developed during pretraining. The existence of such a neural signature across models suggests that self-reflection occupies a distinct and manipulable subspace within the activation manifold.

These visualizations complement our quantitative results and intervention-based analyses, offering geometric intuition for the effectiveness of our vector steering method. Rather than introducing new behaviors from scratch, our method leverages existing latent structures, enhancing or suppressing reflection by operating within naturally emergent manifolds in the model's internal state space.

\section{Identifying Self-Reflection Instances}\label{sec:identify_sr}

To systematically identify self-reflection in language model outputs, we developed a keyword-based detection approach. We define a self-reflection instance as any generation containing explicit self-reflection tokens that signal the model's reconsideration or revision of its reasoning process.

We construct a curated list of self-reflection keywords, informed by prior analyses of reasoning dynamics in language models \cite{guo2025deepseek, liu2024oats}. A generation is marked as self-reflective if it contains one or more of the following terms:

\begin{figure}[h]
\centering
\begin{minipage}[t]{0.48\textwidth}
\vspace{0pt}
\noindent\fbox{%
\parbox{0.95\linewidth}{%
\textbf{Self-Reflection Keywords:}

\smallskip

\texttt{wait}, \texttt{re-check}, \texttt{recheck}, \texttt{check again}, \texttt{rethink}, \texttt{re-think}, \texttt{reconsider}, \texttt{re-consider}, \texttt{try again}, \texttt{re-examine}, \texttt{reexamine}, \texttt{re-evaluate}, \texttt{reevaluate}, \texttt{think again}, \texttt{consider again}, \texttt{evaluate again}, \texttt{examine again}
}%
}
\end{minipage}
\hfill
\begin{minipage}[t]{0.48\textwidth}
\vspace{0pt}
\centering
\includegraphics[width=\linewidth]{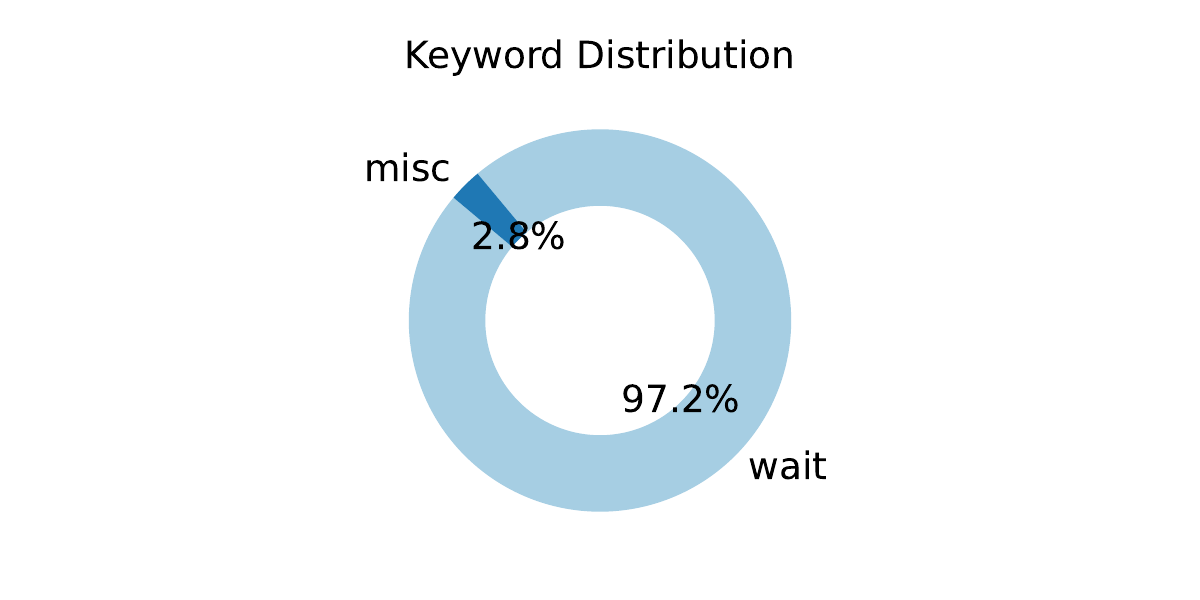}
\label{fig:keyword_freq}
\end{minipage}
\caption{\textbf{Left:} List of self-reflection keywords used in our analysis. \textbf{Right:} Frequency distribution of these keywords across model-generated outputs.}
\end{figure}

To validate this detection method, we applied it to model outputs on the MATH500 benchmark using the DeepSeek-R1-Distill-Qwen-1.5B model. We set the maximum response length to 32{,}784 tokens to accommodate complex, multi-step solutions and to ensure that instances of late-stage self-reflection were not truncated. Among the reflection markers, the keyword \texttt{"wait"} emerged as particularly salient. In our analysis of DeepSeek-R1 outputs, \texttt{"wait"} accounted for approximately 97.2\% of all detected reflection instances. This high frequency makes it a reliable and informative indicator for tracking self-reflective behavior, especially in models fine-tuned with RLVR techniques.

\section{Ablation Study}
\label{app:abl}

We determine the optimal injection strategy via a two-stage procedure:
\begin{enumerate}[label=(\roman*),leftmargin=*]
  \item \textbf{Scaling Search:} For each candidate layer $\ell$, we perform a grid search over $\alpha \in [-1.0, 1.0]$ to identify the value that maximizes validation performance, exploring both enhancement and suppression regimes.

  \item \textbf{Layer Selection:} We evaluate each layer's receptivity to injection. For specialized reasoning models (e.g., DeepSeek-R1), a single well-chosen layer often suffices to yield significant gains. In contrast, for general pretrained models (e.g., Qwen2.5), we observe that distributing moderate injections across multiple layers produces the best trade-off between accuracy and efficiency.
\end{enumerate}

\begin{wrapfigure}{r}{0.48\textwidth}
   \centering
   \includegraphics[width=\linewidth]{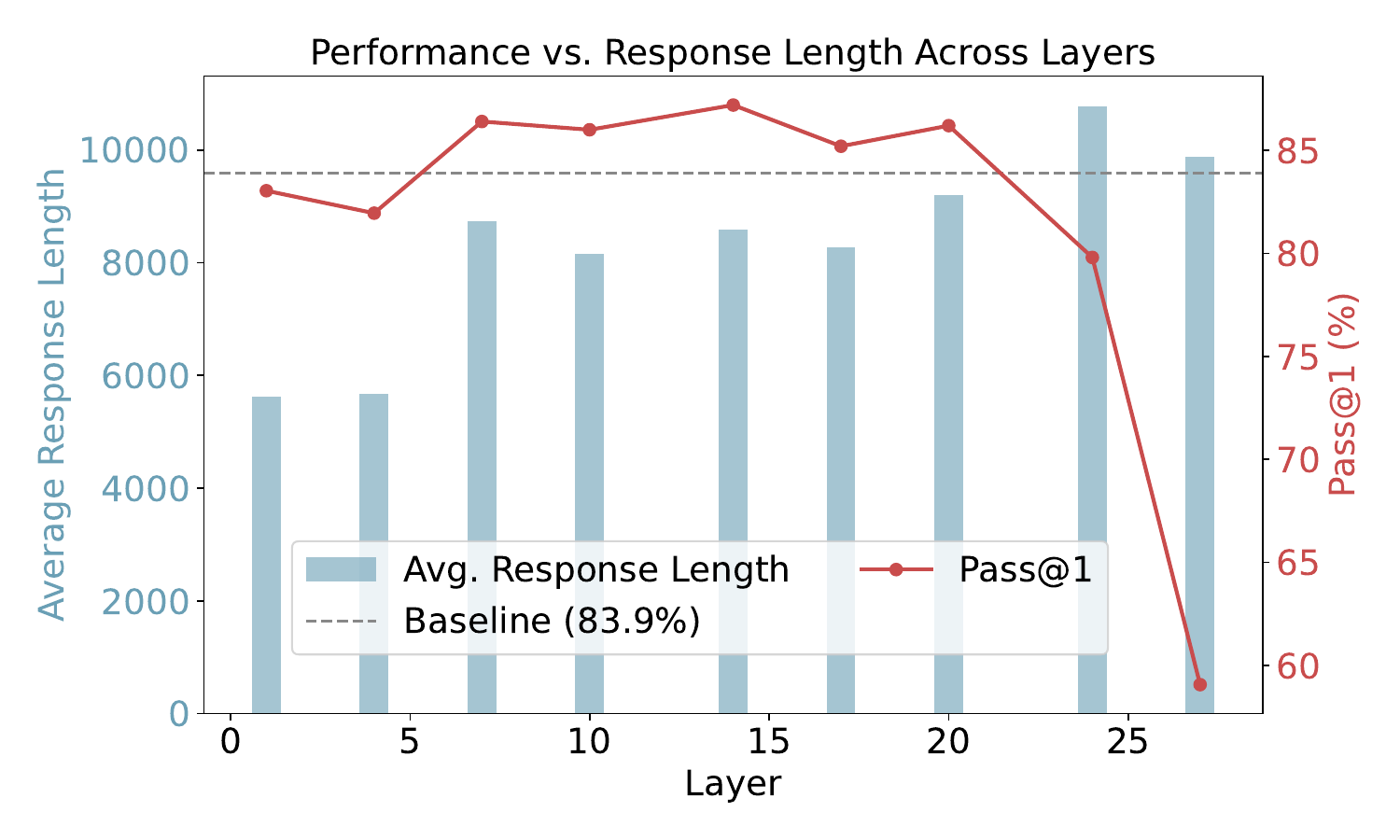}
\caption{Effect of injecting the self-reflection vector at different layers of DeepSeek‑R1 1.5B ($\alpha = 0.01$) on Pass@1 and response length for MATH‑500.}
\vspace{-.1in}
    \label{fig:layer_selection}
\end{wrapfigure}

\paragraph{Effect of Injection Layer.}
We’ve already described $\alpha$ selection in the main text. Here, we fix $\alpha = 0.01$ and examine the effect of injecting the self‑reflection vector at each layer of DeepSeek‑R1 1.5B on MATH‑500 (Figure \ref{fig:layer_selection}).
We observe that middle layers, most notably layer 14, achieve the highest performance. Injections into early layers yield only marginal gains, as the steering signal is progressively transformed and attenuated by subsequent network operations. Conversely, injecting too late often degrades performance, likely because the intervention interferes directly with token generation rather than shaping deeper reasonin prg dynamics. These results indicate that moderate, mid‑network interventions best modulate self‑reflection by targeting layers that both abstract reasoning patterns and retain strong control over final predictions.

\section{Implementation Details} \label{app:imp}
For the DeepSeek-R1-Distill-Qwen models, we adopted a specialized prompting strategy that incorporates the explicit token \texttt{<think>} to elicit self-reflective reasoning and promote internal deliberation.

In contrast, for the Qwen models, we employed the same prompt template but omitted the \texttt{<think>} token. This design allowed us to isolate and assess the specific influence of \texttt{<think>} on eliciting reflective behaviors and its downstream impact, as Qwen models do not natively rely on such explicit triggers.

To support complex multi-step reasoning, we set the maximum generation length to 32,784 tokens across all experiments, ensuring that outputs were not prematurely truncated. All experiments were conducted on a computing cluster equipped with 8 NVIDIA A5000 GPUs.
\section{Limitations.} \label{app:limitation}
Our work presents several limitations.
First, users must predefine whether to enhance or suppress self-reflection prior to inference; the model does not yet autonomously adjust its reflective behavior based on task complexity or reasoning demands.
Second, our approach relies on access to internal model activations, which may not be feasible in closed-source or API-limited environments.

In the future, these limitations could be addressed by developing adaptive self-reflection mechanisms that dynamically modulate introspection based on task complexity and reasoning signals. Further research might extend these techniques to more opaque model environments with limited activation access. Additional work could also explore methods for automatic calibration of injection parameters across diverse model architectures and reasoning domains.

\section{Broader Impacts}\label{app:impacts}

This work focuses on understanding and controlling self-reflection in language models, which carries several potential societal implications. On the positive side, enabling more precise control over the reflection mechanisms in language models can lead to more reliable automated reasoning in high-stakes domains such as medical diagnosis, scientific research, and educational applications. By improving reasoning accuracy through reflection enhancement while maintaining computational efficiency, our technique could make high-quality AI assistance more accessible across resource-constrained environments.

However, we also recognize potential negative impacts. Enhanced reasoning capabilities could be misused for generating more convincing misinformation or deceptive content. Additionally, while our method aims to improve model efficiency, it could potentially increase the environmental impact of AI systems by encouraging more computation-intensive applications that rely on repeated reflection-enhanced reasoning.

Our approach requires direct access to model weights and hidden states, which represents a safeguard against casual misuse, as most harmful applications would be limited to those with substantial technical expertise and computational resources. Furthermore, we aim to mitigate risks by openly sharing our findings with the research community and focusing on techniques that can be used to make models more efficient rather than simply more powerful.

The experiments conducted in this research were limited to benchmark evaluations using publicly available datasets and did not involve human subjects or require IRB approval. Our work did not collect any new data from individuals, nor did it involve deployment in real-world settings that could directly impact people.

\end{document}